\documentclass{article} %
\usepackage{colm2024_conference}

\usepackage{booktabs}
\usepackage{graphicx}
\usepackage{enumitem}
\usepackage{wrapfig}
\usepackage{algorithm}
\usepackage{algpseudocode}

\usepackage{microtype}
\usepackage{multicol}
\usepackage{float}
\usepackage{amsmath}
\usepackage{colortbl}
\usepackage[utf8]{inputenc}
\definecolor{lightgray}{rgb}{0.9,0.9,0.9}
\usepackage{caption}
\usepackage{subcaption}
\usepackage{xcolor}
\usepackage{setspace}
\usepackage{url}
\usepackage{multirow}
\usepackage{colortbl}
\usepackage{tabularx}
\usepackage{blindtext}
\usepackage{pgfplots}
\pgfplotsset{compat=1.18} 
\usepackage{tikz}
\usetikzlibrary{er,positioning,bayesnet}
\usepackage{makecell}
\usepackage{tipa}
\usepackage{siunitx}
\usepackage{nicefrac}
\usepackage{tocloft}
\usepackage{listings}
\usepackage[raster,skins]{tcolorbox} %
\usepackage{xltabular}
\usepackage{adjustbox}
\usepackage{xurl}

\usepackage{amsmath,amsfonts,bm}

\def\eqref#1{equation~\ref{#1}}

\def\1{\bm{1}}

\DeclareMathAlphabet{\mathsfit}{\encodingdefault}{\sfdefault}{m}{sl}
\SetMathAlphabet{\mathsfit}{bold}{\encodingdefault}{\sfdefault}{bx}{n}

\newcommand{\redmoe}{\texttt{dots.llm1}\xspace}
\newcommand{\redmoechat}{dots.llm1.inst\xspace}
\newcommand{\dsvii}{DeepSeek-V2\xspace}

\newcommand{\actparam}{14 billion\xspace}
\newcommand{\totalparam}{142 billion\xspace}

\title{\Large \redmoe Technical Report}

\author{
\bf \large rednote-hilab
}

\begin{document}

\maketitle

\vspace{2cm}

\begin{abstract}
Mixture of Experts (MoE) models have emerged as a promising paradigm for scaling language models efficiently by activating only a subset of parameters for each input token.
In this report, we present \redmoe, a large-scale MoE model that activates \actparam parameters out of a total of \totalparam parameters, delivering performance on par with state-of-the-art models while reducing training and inference costs. 
Leveraging our meticulously crafted and efficient data processing pipeline, \redmoe achieves performance comparable to Qwen2.5-72B after pretraining on 11.2T high-quality tokens and post-training to fully unlock its capabilities. Notably, no synthetic data is used during pretraining. To foster further research, we open-source intermediate training checkpoints at every one trillion tokens, providing valuable insights into the learning dynamics of large language models.

\end{abstract}

\vspace{2cm}

\begin{figure}[hbp]
    \centering
    \includegraphics[width=0.92\textwidth]{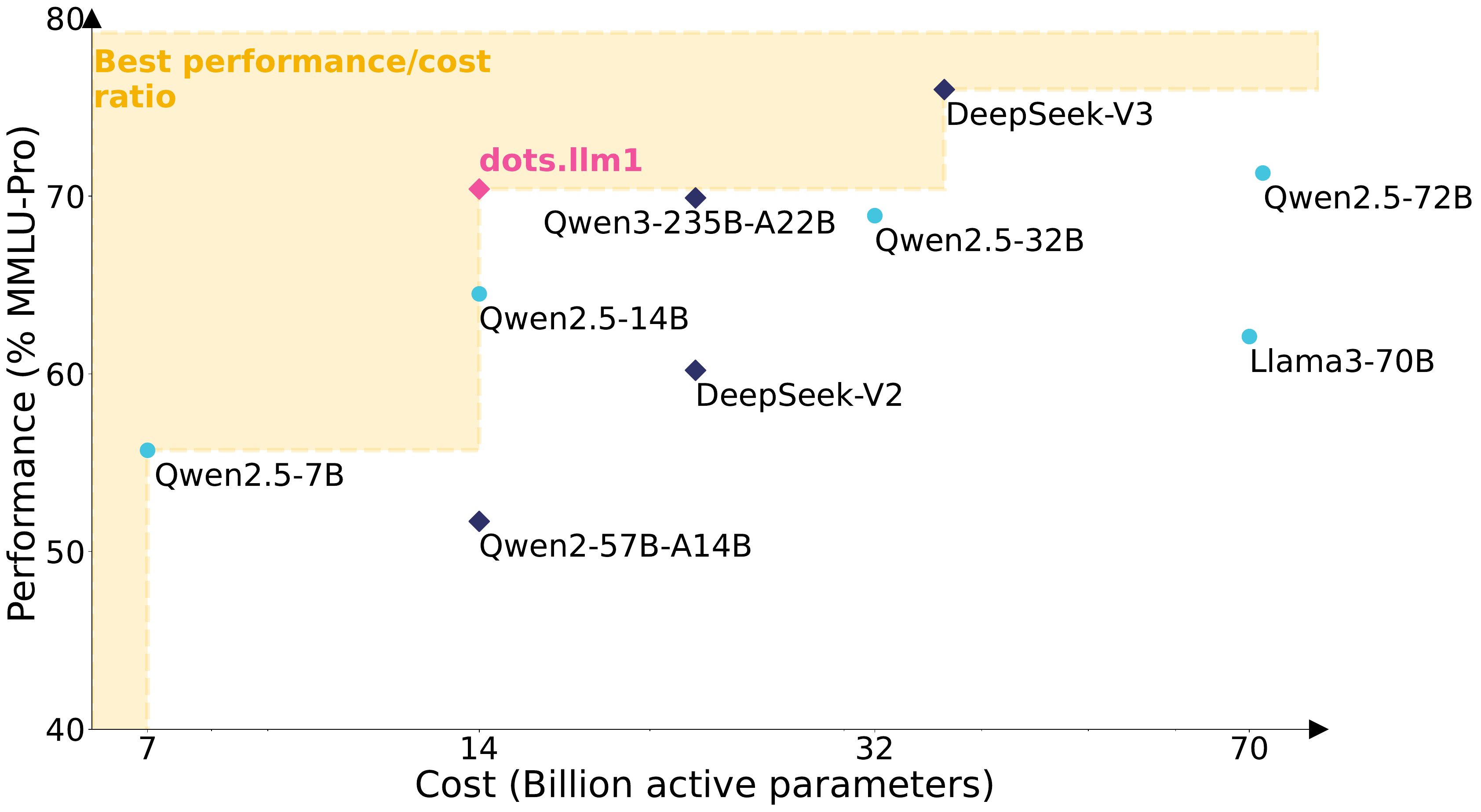}
    \caption{Performance and cost comparison of open MoE and dense language models. Circles ($\circ$) denote dense models, while diamonds ($\diamond$) denote MoE models. We benchmark model capabilities using MMLU-Pro, showing that \redmoe achieves comparable accuracy to leading models.}
    \label{fig:intro}
\end{figure}


\newpage
\section{Introduction}
\label{sec:intro}

Large Language Models (LLMs) have undergone rapid advancements in recent years, moving closer to the goal of Artificial General Intelligence (AGI) as evidenced by substantial progress \citep{gpt45, gpto3mini, claude37, grok3}. Parallel to these proprietary developments, the open-source community is also achieving remarkable breakthroughs \citep{qwen2_5, dsviii, mistralsmall3, llama4}. These initiatives are making substantial efforts to close the performance gap with closed-source models, driving forward the evolution of LLMs in the open-source domain.

Mixture of Experts (MoE) \citep{switch_transformer} is a neural network architecture that divides the model into multiple expert networks, each specializing in different aspects of the input data. By dynamically routing input tokens to a subset of these experts, MoE achieves both computational efficiency and scalability. In recent years, MoE has been widely adopted in the development of large language models (LLMs) \citep{mixtral, qwen_moe, dsviii}, enabling them to scale to large-scale model sizes while maintaining efficient resource utilization. This approach has proven instrumental in pushing the boundaries of LLM performance, as it allows for the activation of only a fraction of the model's parameters during inference, significantly reducing computational overhead. Beyond LLMs, MoE has also found extensive applications in other domains~\citep{riquelme2021scaling, shi2024time, jawaid2024style}, which showcases its versatility and effectiveness.

In this paper, we introduce \redmoe, a powerful and cost-effective mixture-of-experts model with 142B parameters, of which 14B are activated for each token. \redmoe achieves efficient inference on a single node equipped with eight GPUs (40GB or 80GB memory), delivering performance on par with leading open-source models across a wide array of tasks.

The \redmoe model adopts a sparse DeepSeekMoE framework \citep{deepseekmoe}, employing a classic multi-head attention (MHA) \citep{transformer} mechanism combined with QK-Norm to ensure training stability. Additionally, it incorporates the innovative auxiliary-loss-free strategy \citep{noaux_tc, dsviii} to effectively manage load balancing, thereby minimizing any potential negative impacts of the balancing process on model performance.
On the data side, we introduce a three-stage data processing framework to address the pressures of both growing data volume and high-quality requirements through document preparation, rule-based processing, and model-based processing.
On the infrastructure side, we introduce an innovative MoE all-to-all communication and computation overlapping recipe based on \texttt{1F1B} pipeline scheduling and an efficient grouped GEMM implementation to boost computational efficiency.

During the pre-training phase, \redmoe utilizes $11.2$ trillion diverse, high-quality tokens. Following this, the context length is further extended from 8K to 32K. In the post-training phase, the model undergoes efficient supervised fine-tuning (SFT) using 400K carefully curated instruction-tuning instances. \redmoe is evaluated on a suite of pre-training and post-training benchmarks and compared with state-of-the-art models. It demonstrates balanced, robust performance across multiple domains, with notable strengths in Chinese language processing and mathematical reasoning. 

Lastly, we will open-source intermediate training checkpoints at every one trillion tokens, aiming to push the boundaries of LLM development and empower the broader community to build efficient and powerful language models.

We summarize our main contributions as follows:
\begin{itemize}
    \item \textbf{Enhanced Data Processing}: We propose a scalable and fine-grained three-stage data processing framework designed to generate large-scale, high-quality and diverse data for pretraining. The complete process is openly provided to facilitate reproducibility.
    \item \textbf{Performance and Cost Efficiency}: We introduce \redmoe, an open-source model that activates only 14B parameters during inference while delivering comprehensive and computationally efficient performance. Trained on 11.2 trillion high-quality tokens generated through our scalable data processing framework, \redmoe demonstrates robust performance across diverse tasks, all achieved without reliance on synthetic data or model distillation.

    \item \textbf{Infrastructure}: we introduce an innovative MoE all-to-all communication and computation overlapping recipe based on \texttt{1F1B} pipeline scheduling and an efficient grouped GEMM implementation to boost computational efficiency.
    
    \item \textbf{Open Accessibility to Model Dynamics}: By releasing intermediate training checkpoints as open-source, we aim to empower the research community with transparency into the training process, enabling deeper insights into the dynamics of large models and fostering accelerated innovation in the field of LLM.


\end{itemize}

\section{Architecture }

The \redmoe model is a decoder-only Transformer architecture \citep{transformer}, where each layer consists of an attention layer and a Feed-forward Network(FFN). 
Unlike dense models such as Llama \citep{llama3_model_card} or Qwen \citep{qwen2.5}, the FFN is replaced with a Mixture of Experts (MoE) module \citep{mixtral, dsvi, dsvii, dsviii}. This modification allows for the training of highly capable models while maintaining an economical cost.


\paragraph{Attention Layer} We utilize a vanilla multi-head attention mechanism~\citep{transformer} in our model. Following \cite{dehghani2023scaling}, RMSNorm is applied to the query and key projections prior to computing attention. This normalization mitigates the risk of excessively large attention logits, which could otherwise destabilize the training process~\citep{wortsman2023small, tian2024nyonic, olmo20242}.

\paragraph{Mixture-of-Experts Layer} Following \cite{dsvi, qwen_moe}, we replace the FFN with a Mixture-of-Experts (MoE) layer comprising both shared and isolated experts. Our implementation features 128 routed experts and 2 shared experts activated for all tokens, with each expert implemented as a fine-grained, two-layer FFN utilizing \texttt{SwiGLU}~\citep{shazeer2020glu} activation. For each token, the router selects the top-6 isolated experts in addition to the 2 shared experts, resulting in 8 active experts per token. Notably, we employ FP32 precision for the gating layer computations rather than BF16 to ensure numerical stability and more accurate expert selection during the routing process.

\paragraph{Load Balancing} Imbalanced expert loading in the MoE layer can lead to routing collapse, diminishing both model capacity and computational efficiency during training and inference.
To address this issue, we adopt an auxiliary-loss-free approach~\citep{noaux_tc}, as also employed in~\cite{dsviii}. It introduces a bias term for each expert, which is added to the corresponding affinity scores to determine the top-k routing. This bias term is dynamically adjusted during training to maintain a balanced load across experts. In addition, we also employ a sequence-wise balance loss to prevent extreme imbalance within any single sequence. Due to the effective load balancing strategy, \redmoe maintains good load balance throughout training and does not drop any tokens during training.

\section{Infrastructures}

The training of \redmoe is underpinned by the internal Cybertron framework, which is a lightweight training framework built upon Megatron-Core.  Leveraging Megatron-Core, we have meticulously constructed a comprehensive suite of toolkits for both model pre-training and post-training. For distinct training stages, namely pre-training, supervised fine-tuning (SFT), and reinforcement learning (RL), we have encapsulated separate trainers in a unified manner to guarantee the coherence and high efficiency of the training process.

\subsection{Interleaved \texttt{1F1B} based Communication and Computation Overlap}

We propose an innovative interleaved \texttt{1F1B} based all-to-all communication and computation overlap solution~\citep{ovelap_1f1b}, and work with NVIDIA to integrate it into Megatron-Core.
Compared to the vanilla interleaved \texttt{1F1B} pipeline, we add an additional step to the warm-up phase, which incurs neither additional bubble overheads nor additional GPU memory consumption for activations.
Under this refined pipeline scheduling scheme, during the steady \texttt{1F1B} phase, all-to-all communication and computation within the forward and backward step pairs can be effectively overlapped, similar to the DualPipe proposed by DeepSeek~\citep{dsviii}.
Compared with the DualPipe, our approach demonstrates a notable advantage in terms of memory consumption, albeit with a marginally higher bubble rate.

\begin{figure*}[t]
    \centering
    \includegraphics[width=0.95\textwidth]{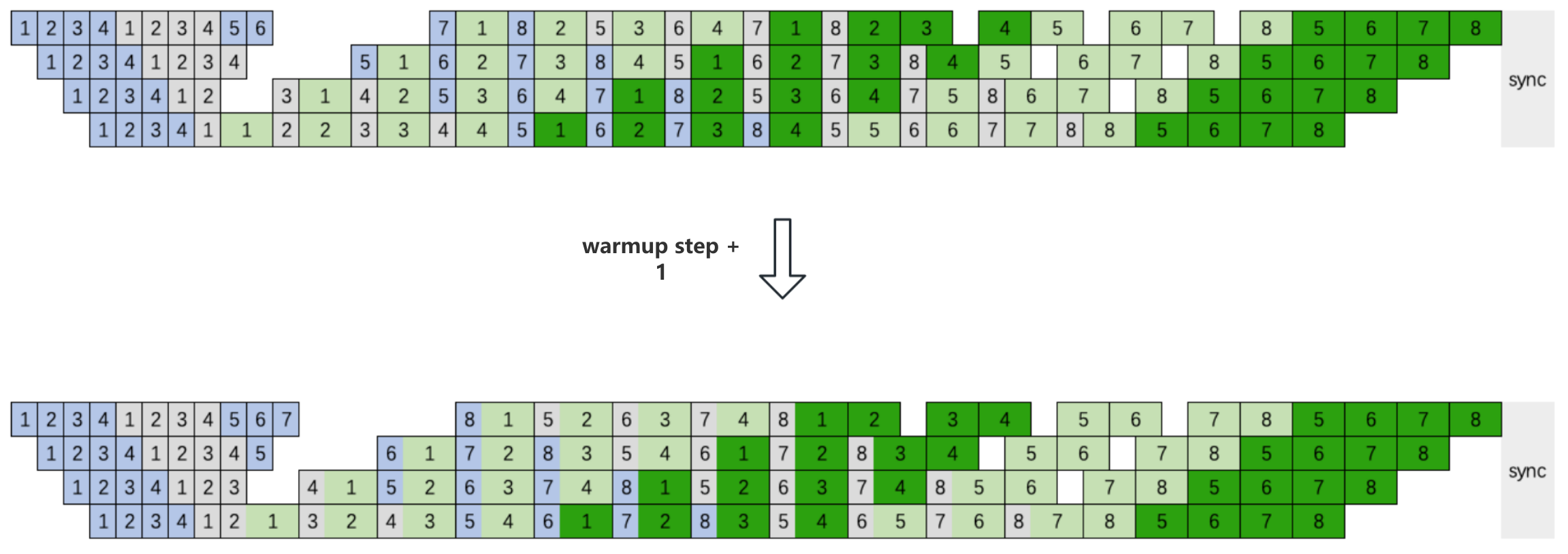}
    \caption{Interleaved \text{1F1B} based communication and computation overlap}
    \label{fig:utk_main}
\end{figure*}

\subsection{Efficient Implementation of Grouped GEMM}

Grouped GEMMs play a significant role in the computation of MoE architectures. A straightforward approach is to schedule sub-GEMM problems in a unified manner across streaming multiprocessors. Some frameworks pre-compute the mapping between tiles and thread blocks on the host side to reduce scheduling overhead on the device~\citep{DBLP:conf/ppopp/Li0YJL19}. Other works have explored scheduling strategies for small or variable-sized batched GEMMs~\citep{DBLP:journals/tjs/YangLW22, DBLP:conf/hpcc/ZhangWMZSXXY22}. Additionally, NVIDIA has introduced Grouped GEMM APIs in cuBLAS~\citep{Nvidia2024GroupedGEMM} and Transformer Engine (TE)~\citep{Nvidia2024TEGroupedGEMM}.

To allow large tile sizes and reduce scheduling overhead, we align $M_i$ (token segment of expert $i$) to a fixed block size.
This fixed block size must be a multiple of \texttt{M} in tile shape modifier \texttt{mMnNkK} of the asynchronous warpgroup level matrix multiply-accumulate (WGMMA) (\texttt{wgmma.mma\_async}) instruction.
Thus, the warpgroups in a single threadblock will have a unified tiling, and the whole token segment ($M_i$) processed by a threadblock necessarily belongs to the same expert, making the scheduling very similar to that of normal GEMMs.

\begin{table}[t]
\centering
\caption{Performance comparison of Transformer Engine 2.1 and our implementation on H800. Here, $g$, $m$, $n$, and $k$ represent the number of groups (experts) and the dimensions of each sub-GEMM problem, respectively.}
\label{tab:infra_grouped_gemm_perf}
\setlength{\tabcolsep}{16pt}
\begin{tabular}{ccccc}
\toprule
& (g, m, n, k) & \makecell{TE's\\(TFLOPS)} & \makecell{Ours\\(TFLOPS)} & Speed-up \\
\midrule
\multirow{8}{*}{forward}  & (4, 1024, 2816, 4096) & 589.12 & 662.48 & 12.45\% \\
                          & (4, 1024, 4096, 2816) & 581.77 & 725.35 & 24.68\% \\
                          & (4, 2048, 2816, 4096) & 636.49 & 678.89 &  6.66\% \\
                          & (4, 2048, 4096, 2816) & 636.94 & 745.45 & 17.04\% \\
                          & (8, 1024, 2816, 4096) & 614.87 & 654.77 &  6.49\% \\
                          & (8, 1024, 4096, 2816) & 609.69 & 745.01 & 22.19\% \\
                          & (8, 2048, 2816, 4096) & 646.70 & 705.85 &  9.15\% \\
                          & (8, 2048, 4096, 2816) & 657.20 & 744.74 & 13.32\% \\
\midrule
\multirow{8}{*}{backward} & (4, 1024, 2816, 4096) & 614.24 & 665.06 & 8.27\% \\
                          & (4, 1024, 4096, 2816) & 594.03 & 637.48 & 7.31\% \\
                          & (4, 2048, 2816, 4096) & 655.39 & 709.75 & 8.29\% \\
                          & (4, 2048, 4096, 2816) & 635.02 & 677.51 & 6.69\% \\
                          & (8, 1024, 2816, 4096) & 633.35 & 677.56 & 6.98\% \\
                          & (8, 1024, 4096, 2816) & 619.97 & 637.71 & 2.86\% \\
                          & (8, 2048, 2816, 4096) & 666.00 & 706.85 & 6.13\% \\
                          & (8, 2048, 4096, 2816) & 650.53 & 695.38 & 6.89\% \\
\bottomrule
\end{tabular}
\end{table}

Compared to NVIDIA’s Grouped GEMM APIs in Transformer Engine (v2.1), our implementation exhibits notable advantages. Table~\ref{tab:infra_grouped_gemm_perf} presents a performance comparison for forward and backward computations on the H800, where tokens are evenly routed to experts. Our approach achieves an average performance improvement of 14.00\% for forward computation and 6.68\% for backward computation.

\section{Pre-training}

\subsection{Pre-training Data}

The quantity, quality, and diversity of training data play a crucial role in determining the performance of language models. To achieve these objectives, we have structured our approach into three key stages: document preparation, rule-based processing, and model-based processing. This design allows us to efficiently manage large volumes of data, even with limited computational resources. 
Document preparation focuses on preprocessing and organizing the raw data.
Rule-based processing aims to minimize the need for extensive human curation by automatically filtering and cleaning the data.
Model-based processing further ensures that the final dataset is both high-quality and diverse.
To ensure safety, we applied filters to exclude data from websites likely to contain unsafe content or significant amounts of personally identifiable information (PII), as well as domains flagged as harmful by a safety classifier.
During pre-training, we maintain a balanced 1:1 ratio of Chinese to English data, and no synthetic data is used.

The data processing pipeline is thoroughly documented in \autoref{appsec:data}. Two key innovations distinguish our data processing pipeline as follows:

\paragraph{Web Clutter Removal Model} To address issues such as boilerplate content and repetitive lines, we develop a lightweight model that operates at the line level. This approach achieves an effective balance between cleaning quality and computational efficiency, representing a unique feature that is not commonly found in open-source datasets.

\paragraph{Category Balancing} We train a 200-class classifier to balance the proportions within the web data. This enables us to increase the presence of knowledge-based and factual content, such as encyclopedia entries and popular science articles, while reducing the share of fictional and highly structured web content, including science fiction novels and product descriptions.






\subsection{Hyper-Parameters}

The \redmoe model is trained using the AdamW optimizer \citep{adamw}, with $\beta_1 = 0.9$ and $\beta_2 = 0.95$. We implement a weight decay of $0.1$ and apply gradient clipping at $1.0$. Following \cite{deepseekmoe, dsvii, dsviii}, the weights are initialized from a normal distribution with standard deviation 0.006. \redmoe comprises 62 layers, with the first layer utilizing a vanilla dense FFN and the subsequent layers employing MoE.

We set the maximum sequence length to 8K during pre-training and train \redmoe on 11.2T tokens. Following \cite{minicpm}, we utilize a warmup-stable-decay learning rate schedule. The learning rate warms up over 4,000 steps before stabilizing at $3 \times 10^{-4}$ for the stable training phase, which encompasses 10T tokens of data. We progressively increase the batch size from 64M tokens initially to 96M tokens at 6T tokens, and finally to 128M tokens at 8.3T tokens.

After the main training phase, the process includes two annealing stages, comprising a total of 1.2 trillion tokens of data.

Stage 1: We train for 1T tokens while gradually decreasing the learning rate from $3 \times 10^{-4}$ to $3 \times 10^{-5}$. During this annealing stage, we significantly increase the proportion of reasoning-related and knowledge-related data to 90\%.

Stage 2: We continue training for 200B tokens while reducing the learning rate from $3 \times 10^{-5}$ to $1 \times 10^{-5}$ and increasing the proportion of code, mathematics, and reasoning data.

\subsection{Long Context Extension}

We implement context length extension after the annealing phases. During this phase, we maintain a constant learning rate while training on 128B tokens using UtK strategy~\citep{utk}, extending the sequence length to 32K. Rather than modifying the datasets, UtK tries chunking of training documents into smaller segments, then training the model to reconstruct relevant segments from shuffled chunks. By learning to untie these knotted chunks, the model can effectively handle longer input sequences while maintaining its performance on short-context tasks.

\subsection{Evaluation}

\subsubsection{Benchmarks}

To comprehensively evaluate \redmoe model, which is pretrained on both Chinese and English, we assess its performance across a suite of benchmarks spanning multiple domains in each language. All evaluations are conducted using the vLLM framework~\citep{vllm}. We categorize the benchmarks as follows:

\paragraph{Language Understanding} To evaluate English contextual reasoning and reading comprehension, we employ benchmarks such as HellaSwag~\citep{hellaswag}, PIQA~\citep{piqa}, the ARC suite~\citep{arc}, BigBenchHard (BBH)~\citep{bbh}, and DROP~\citep{drop}. For English closed-book question answering, we utilize TriviaQA~\citep{joshi-etal-2017-triviaqa} and Natural Questions~\citep{naturalquestions}. For Chinese language understanding, we assess reference disambiguation using CLUEWSC~\citep{clue} and reading comprehension using C3~\citep{sun2019investigating}.

\paragraph{Knowledge} To measure domain knowledge in English, we leverage the MMLU~\citep{mmlu}, MMLU-Pro~\citep{mmlu_pro} and SuperGPQA~\citep{supergpqa} benchmarks as well as  for creative problem-solving and AGIEval~\citep{agieval} for standardized exam performance. For the assessment of Chinese knowledge, we employ C-Eval~\citep{ceval}, CMMLU~\citep{cmmlu}, and Xiezhi~\citep{gu2024xiezhieverupdatingbenchmarkholistic}.

\paragraph{Mathematics} To evaluate mathematical reasoning capabilities, we utilize CMath~\citep{wei2023cmath} and GSM8K~\citep{gsm8k} for foundational problem-solving tasks, as well as MATH~\citep{math} for assessing advanced mathematical problem-solving proficiency.

\paragraph{Code} Code generation skills are evaluated with HumanEval~\citep{humaneval}, MBPP~\citep{mbpp} and MCEval~\citep{mceval}. Additionally, the ability to handle diverse function calls and complex instructions is measured through the BigCodeBench benchmark~\citep{bigcodebench}.

\subsubsection{Results}
\label{subsec:results}

\newcommand{\avgrowcolor}{\cellcolor[HTML]{D7E8E8}}

\begin{table}[t]
    \centering
    \footnotesize
    \setlength{\tabcolsep}{6pt}
    \caption{
        Comparison between \redmoe and other representative open-source base models. All models were evaluated under identical conditions. Results demonstrate that \redmoe achieves superior performance compared to \dsvii{}, while delivering results comparable to Qwen2.5 72B.
    }
    \label{tab:base}
    \begin{tabular}{@{}c l c | c c | c c | c@{}}
    \toprule
 & \multirow{2}{*}{\centering \textbf{Benchmark {\tiny (Metric)}}} & \multirow{2}{*}{\textbf{\# Shots}} & \textbf{Qwen2.5} & \textbf{Qwen2.5} & \textbf{DeepSeek} & \textbf{DeepSeek} & \textbf{\redmoe} \\
     &  &  & \textbf{32B Base} & \textbf{72B Base} & \textbf{V2 Base} & \textbf{V3 Base} & \textbf{\texttt{.base}} \\ 
     \midrule
     & Architecture & - & Dense & Dense & MoE & MoE & MoE \\
     & \# Activated Params & - & 32B & 72B & 21B & 37B & 14B \\
     & \# Total Params & - & 32B & 72B & 236B & 671B & 142B \\ 
     \midrule
    \multirow{6}{*}{Chinese} & C-Eval {\tiny (EM)} & 5-shot & 87.4 & 89.3 & 80.6 & 89.1 & 92.8 \\
     & CMMLU {\tiny (EM)} & 5-shot & 88.5 & 89.7 & 82.9 & 88.1 & 90.4 \\
     & CLUEWSC {\tiny (EM)} & 5-shot & 93.2 & 93.2 & 92.2 & 92.6 & 92.7 \\
     & C3 {\tiny (EM)} & 5-shot & 97.2 & 97.1 & 96.7 & 97.4 & 97.6 \\
     & Xiezhi {\tiny (EM)} & 3-shot & 81.2 & 82.3 & 77.4 & 80.4 & 82.9 \\
     & \avgrowcolor Average & \avgrowcolor & \avgrowcolor 89.5 & \avgrowcolor 90.3 & \avgrowcolor 86.0 & \avgrowcolor 89.5 & \avgrowcolor 91.3 \\ 
     \midrule
    \multirow{12}{*}{English} & TriviaQA {\tiny (EM)} & 5-shot & 78.5 & 85.2 & 89.2 & 91.4 & 88.2 \\
     & BBH {\tiny (EM)} & 3-shot & 82.2 & 83.7 & 76.5 & 83.5 & 80.8 \\
     & MMLU {\tiny (EM)} & 5-shot & 83.5 & 85.4 & 78.0 & 86.8 & 83.2 \\
     & MMLU-Pro {\tiny (EM)} & 5-shot & 61.8 & 64.7 & 52.3 & 61.4 & 61.9 \\
     & SuperGPQA {\tiny (EM)} & 5-shot & 30.5 & 32.1 & 29.0 & 38.6 & 33.3 \\
     & DROP {\tiny (F1)} & 3-shot & 82.9 & 84.9 & 78.9 & 89.1 & 83.4 \\
     & ARC-Challenge {\tiny (EM)} & 25-shot & 93.1 & 95.2 & 92.1 & 96.2 & 93.8 \\
     & AGIEval {\tiny (EM)} & 3-shot & 76.1 & 76.9 & 68.34 & 77.1 & 79.9 \\
     & HellaSwag {\tiny (EM)} & 10-shot & 93.3 & 94.2 & 88.1 & 89.7 & 88.2 \\
     & PIQA {\tiny (EM)} & 5-shot & 92.8 & 94.5 & 90.9 & 94.0 & 91.0 \\
     & NaturalQuestions {\tiny (EM)} & 3-shot & 35.3 & 42.7 & 46.8 & 50.7 & 48.7 \\
     & \avgrowcolor Average & \avgrowcolor & \avgrowcolor73.2 & \avgrowcolor 76.3 & \avgrowcolor 71.8 & \avgrowcolor 78.0 & \avgrowcolor 75.7 \\ 
     \midrule
    \multirow{4}{*}{MATH} & GSM8K {\tiny (EM)} & 8-shot & 88.9 & 89.9 & 80.0 & 90.1 & 86.7 \\
     & MATH {\tiny (EM)} & 4-shot & 71.7 & 76.2 & 42.3 & 72.8 & 68.7 \\
     & CMath {\tiny (EM)} & 3-shot & 79.2 & 65.7 & 72.2 & 83.3 & 79.5 \\
     & \avgrowcolor Average & \avgrowcolor & \avgrowcolor79.9 & \avgrowcolor 77.3 & \avgrowcolor 64.8 & \avgrowcolor 82.1 & \avgrowcolor 78.3 \\ 
     \midrule
    \multirow{5}{*}{Code} & HumanEval {\tiny (Pass@1)} & 0-shot & 50.6 & 56.7 & 43.3 & 62.8 & 64.0 \\
     & MBPP {\tiny (Pass@1)} & 3-shot & 77.0 & 77.8 & 75.1 & 76.3 & 76.7 \\
     & MCEval {\tiny (Pass@1)} & 0-shot & 47.9 & 47.5 & 43.5 & 50.1 & 46.6 \\
     & BigCodeBench {\tiny (Pass@1)} & 0-shot & 51.9 & 54.0 & 41.8 & 60.7 & 50.9 \\
     & \avgrowcolor Average & \avgrowcolor & \avgrowcolor56.9 & \avgrowcolor 59.0 & \avgrowcolor 50.9 & \avgrowcolor 62.5 & \avgrowcolor 59.6 \\ 
    \bottomrule
    \end{tabular}
\end{table}

As shown in \autoref{tab:base}, we compare \redmoe with other leading open-source base models under identical conditions. The results demonstrate that \redmoe, with only 14B activated parameters, achieves superior performance compared to DeepSeek-V2, and delivers results comparable to Qwen2.5 72B. This makes \redmoe one of the most economical and powerful mixture-of-experts models available to date.

For a detailed breakdown across different categories, we use Qwen2.5 72B as the baseline for comparison. \redmoe demonstrates comparable performance to Qwen2.5 72B across most domains. (1) On language understanding tasks, \redmoe achieves superior performance on Chinese language understanding benchmarks, which can be attributed to our data processing pipeline. (2) In knowledge tasks, while \redmoe shows slightly lower scores on English knowledge benchmarks, its performance on Chinese knowledge tasks remains robust. (3) In the code and mathematics domains, \redmoe achieves higher scores on HumanEval and CMath. Interestingly, for mathematics, we observe that \redmoe achieves better performance in the zero-shot setting compared to the few-shot setting, with an improvement of over 4 points. We leave further investigation of this phenomenon to future work.

\subsection{Analysis}

\begin{figure}[t]
    \centering
    \includegraphics[width=0.95\textwidth]{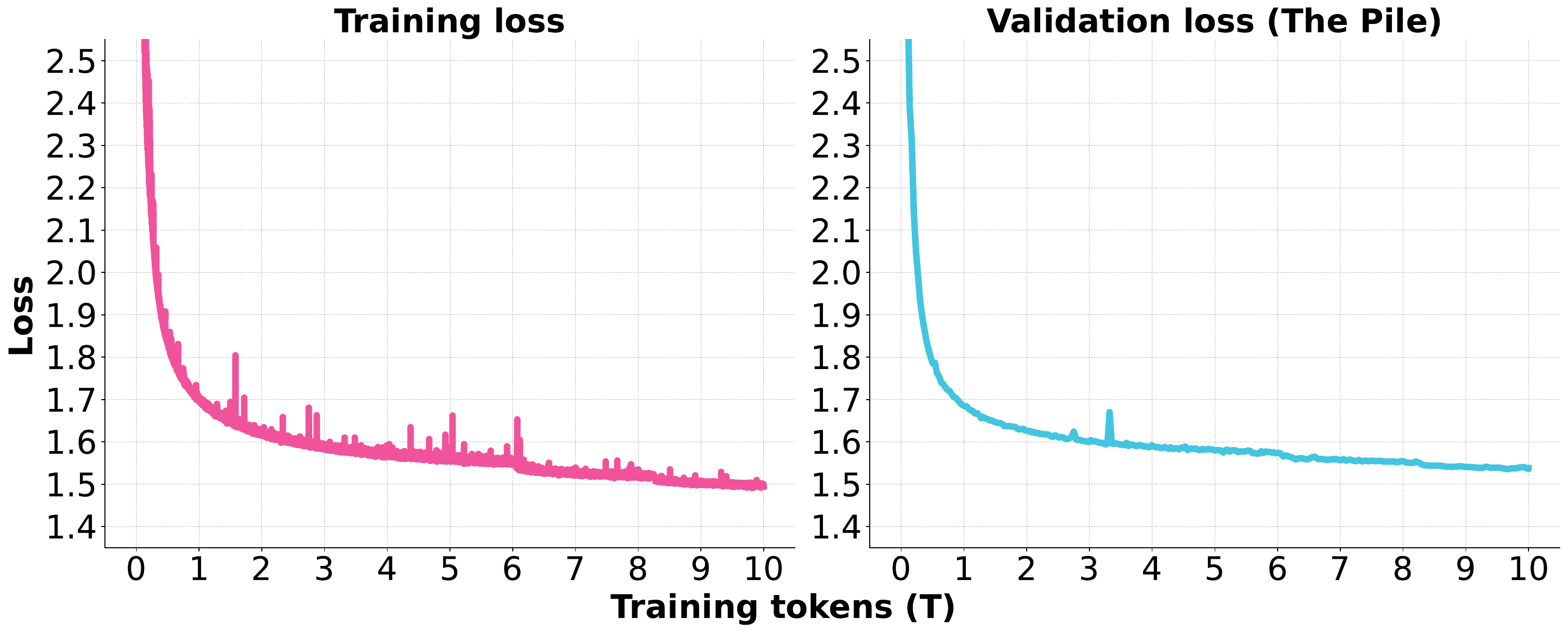}
    \caption{The loss curve highlights the consistent stability of the training process. At 6 trillion training tokens, we adjusted our batch size from 64 million to 96 million. At 8.3 trillion, we further increased it to 128 million.}
    \label{fig:loss}
\end{figure}

\paragraph{Stable Training} As depicted in \autoref{fig:loss}, the loss curve highlights the remarkable stability of the training process. Throughout the training period, there are no incidents of irrecoverable loss spikes or the need for rollback operations.

\paragraph{High Quality Web Data}

\begin{figure}[t]
    \centering
    \includegraphics[width=0.92\textwidth]{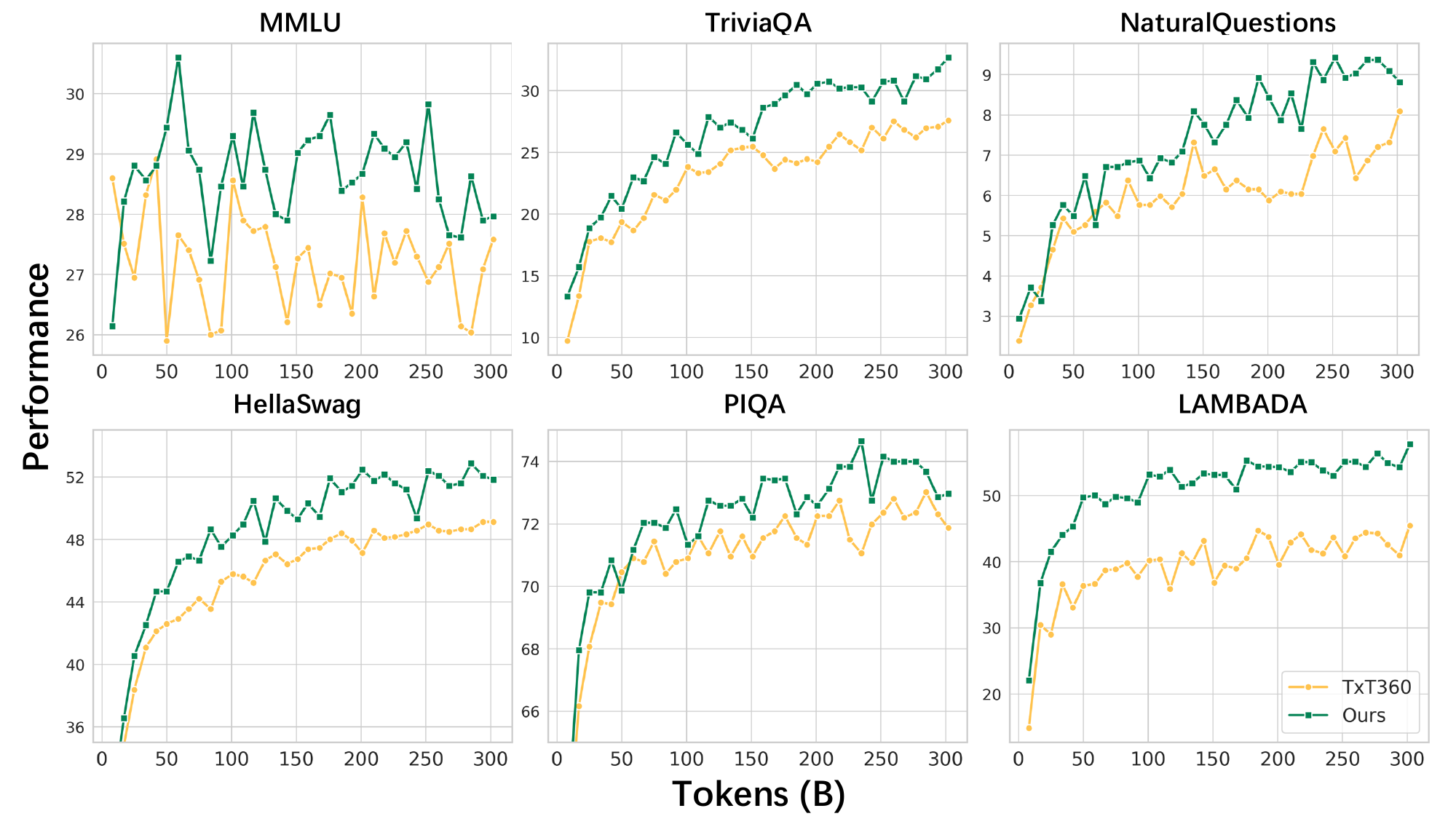}
    \caption{Comparison of performance curves between the TxT360 dataset and our web data on the MMLU, TriviaQA, NaturalQuestions, HellaSwag, PIQA, and LAMBADA~\citep{lambada} benchmarks. The results demonstrate that training with our web data consistently yields superior performance.}
    \label{fig:cc}
\end{figure}

To assess the quality of our web data, we conduct validation experiments comparing it to the TxT360 dataset \citep{txt360data2024}, which represents the current state-of-the-art (SOTA) in open-source web data. The experiments are performed using a 1.5 billion parameter dense model, with the same architecture as Qwen2.5 1.5B. We randomly sample 350 billion tokens for training to obtain reliable experimental results.
As illustrated in \autoref{fig:cc}, our results surpass those of TxT360, demonstrating the high quality of our web data through our processing pipeline.

\paragraph{Effective Long Context Extension}

We use RULER~\citep{hsieh2024ruler} on the base models to evaluate long-context capabilities. As shown in \autoref{tab:ruler}, \redmoe delivers competitive performance at both 8K and 16K context lengths.

\begin{table}[t]
\centering
\setlength{\tabcolsep}{16pt}
\caption{Long Context Performance on the RULER benchmark.}
\label{tab:ruler}
\begin{tabular}{lcccc}
\toprule
Model & 4K & 8K & 16K & 32K \\
\midrule
Qwen2.5 72B & 96.5 & 94.3 & 93.1 & 92.7 \\
\redmoe & 94.7 & 94.9 & 92.6 & 87.7 \\
\bottomrule 
\end{tabular}
\end{table}

\paragraph{Economical Training Cost}

As highlighted in \autoref{tab:cost}, \redmoe showcases significant efficiency in terms of training costs. 
The comparison reveals that our approach substantially reduces the GPU hours required per one trillion tokens. Qwen2.5 72B model (in our optimized framework) requires 340K GPU hours, while \redmoe reduces this to just 130K GPU hours, leading to substantial savings in computational resources and expenses. 
Furthermore, if we consider the entire pretraining process, \redmoe requires only 1.46 million GPU hours, whereas Qwen2.5 72B consumes 6.12 million GPU hours, representing a reduction of 4$\times$ in total computational resources. This substantial gap demonstrates the cost-effectiveness and scalability of \redmoe, making it a more economical choice for large-scale pretraining.

\begin{table}[t]
\centering
\caption{Comparison of GPU hours across different models, \redmoe representing a reduction of $4\times$ in total computational resources.}
\begin{tabular}{lccc}
\toprule
Model & GPU Hours per One Trillion Tokens & Pretraining Tokens & Total GPU Hours \\
\midrule
Qwen2.5 72B & 340K ($\times 1.0$) & 18.0T & 6,120K ($\times 1.0$) \\
\redmoe & 130K ($\times 0.38$) & 11.2T & 1,456K ($\times 0.24$)\\ 
\bottomrule
\end{tabular}
\label{tab:cost}
\end{table}

\section{Post-training}
\label{sec:post}


\subsection{Supervised Fine-tuning}

\paragraph{Data Mixture}
Based on open-source data and internally annotated data, we collect approximately 400K instruction tuning instances, focusing on several key areas: multi-lingual (primarily Chinese and English) multi-turn dialogues, knowledge understanding and question answering, complex instruction following, and reasoning tasks involving mathematics and coding.

For multi-turn dialogue capabilities, we integrate mainstream open-source dialogue datasets with our extensive collection of high-quality internal Chinese instruction sets.
For a small subset of responses in the open-source data that are of insufficient quality, we refine them using powerful teacher models (e.g., DeepSeek-V3 0324 \citep{dsviii}).
To enhance knowledge understanding QA abilities, we specifically incorporate datasets focused on factual knowledge and reading comprehension. For complex instruction following, we meticulously design instruction sets with conditional constraints and filtered out responses that failed to adhere to these conditions. To develop reasoning capabilities, we introduce verifiable mathematics and coding datasets, equipped with corresponding verifiers to filter and extract high-quality supervision signals.

\paragraph{Fine-tuning Configurations}

The fine-tuning process of \redmoechat{} consists of two phases. In the first phase, we perform upsampling and multi-session concatenation on our 400K instruction tuning instances, then fine-tune \redmoechat{} for 2 epochs. 
In the second phase, we further enhance the model's capabilities in specific domains (such as mathematics and coding) through rejection sampling fine-tuning (RFT), incorporating a verifier system to improve performance in these specialized areas.
During each training phase, we employ a cosine learning rate scheduler with a learning rate of 5e-6 that gradually decayed to a minimum of 1e-6.

\subsection{Evaluations}

\paragraph{Benchmarks}

We evaluate \redmoechat{} on a series of post-training benchmarks and compared it with state-of-the-art models~\citep{dsvii, dsviii}. Our evaluation framework covers five main categories: English general capabilities, Chinese general capabilities, alignment, and two specific subdomains—mathematics and coding.

For English general capabilities, we primarily use representative QA datasets, such as general question-answering benchmarks like MMLU~\citep{mmlu}, MMLU-Redux~\citep{mmlu_redux}, and MMLU-Pro~\citep{mmlu_pro}, expert-level question-answering benchmarks like GPQA~\citep{gpqa}, as well as reading comprehension QA datasets such as DROP~\citep{drop}. Additionally, SimpleQA~\citep{simpleqa} evaluates model's ability to answer fact-seeking questions correctly.

For Chinese general capabilities, we select CLUEWSC~\citep{clue} for coreference resolution evaluation, C-Eval~\citep{ceval} for comprehensive Chinese language understanding capabilities, and C-SimpleQA~\citep{he2024chinese} for factual question answering in Chinese.

To measure the instruction-following and human preference alignment of the \redmoechat{} model, we select rule-based verifiable benchmarks (IFEval~\citep{zhou2023instruction}) and competitive arena-style benchmarks (AlpacaEval2~\citep{dubois2024length}, ArenaHard~\citep{li2024crowdsourced}).

For mathematics, we evaluate on competition-level benchmarks AIME24~\citep{AIME2024} and CNMO24~\citep{liu2024your}, and standard-difficulty benchmarks MATH~\citep{math}, GSM8K~\citep{gsm8k}, and MATH500 \citep{lightman2023let}.

For coding capabilities, we conduct tests on the coding competition LiveCodeBench~\citep{jain2024livecodebench}, as well as HumanEval~\citep{humaneval} and MBPP+~\citep{liu2024evaluating}.

\paragraph{Evaluation Configurations}

For general question-answering benchmarks (including MMLU-Pro, DROP, GPQA, SimpleQA, and Chinese SimpleQA), we utilize evaluation prompts provided by the simple-evals\footnote{\url{https://github.com/openai/simple-evals}} framework, while for MMLU-Redux, we apply the zero-shot prompt template from the Zero-Eval\footnote{\url{https://github.com/WildEval/ZeroEval}} repository. For other general and alignment datasets, we follow the evaluation methodologies designed by the original dataset creators.

In mathematics evaluations, we adhere to standard practices for mathematical datasets, using ``Please reason step by step, and put your final answer within \verb|\boxed{}|" for answer extraction. For competitive mathematics benchmarks (e.g., AIME24), we employ temperature sampling and multiple-run averaging to enhance evaluation stability.
In our code evaluations for LiveCodeBench, we select data spanning from May 2024 to January 2025. For evaluating MBPP+, we utilized the v0.2.0 release of EvalPlus~\citep{liu2023your}, which provides reliable assessment for code generation tasks.

\begin{table}[t]
    \centering
    {\fontsize{8.5pt}{9pt}\selectfont  
    \setlength{\tabcolsep}{2pt}
    \caption{Comparison between \redmoechat{} and other representative instruction-tuned models. All models were evaluated within our internal framework under identical conditions. Regarding the Qwen3 series, our evaluations were conducted without the thinking mode enabled. 
    }
    \label{tab:main}
    \begin{tabular}{@{}c l | c c | c c | c c | c | c@{}}
    \toprule
    & \multirow{2}{*}{\centering \textbf{Benchmark {\tiny (Metric)}}}  & \textbf{Qwen-2.5} & \textbf{Qwen-2.5} & \textbf{Qwen-3} & \textbf{Qwen-3} & \textbf{DeepSeek} & \textbf{DeepSeek} & \textbf{gpt4o} & \textbf{\redmoe} \\
    & & \textbf{32B Inst} & \textbf{72B Inst} & \textbf{32B} & \textbf{235B-A22B} & \textbf{V2 Chat} & \textbf{V3} & 
 \textbf{0806} & \textbf{\texttt{.inst}} \\
    \midrule
    & Architecture & Dense & Dense & Dense & MoE & MoE & MoE & - & MoE \\
    & \# Activated Params & 32B & 72B & 32B & 22B & 21B & 37B & - & ~14B \\
    & \# Total Params & 32B & 72B & 32B & 235B & 236B & 671B & - & ~142B \\
    \midrule
   \multirow{7}{*}{English} & MMLU {\tiny (EM)} & 83.4 & 84.4 & 82.9 & 86.4 & 78.6 & 87.9 & 86.7 & 82.1 \\
    & MMLU-Redux {\tiny (EM)} & 83.3 & 85.9 & 86.1 & 88.6 & 77.9 & 88.8 & 87.3 & 85.1 \\
    & MMLU-Pro {\tiny (EM)} & 68.9 & 71.0 & 68.3 & 69.9 & 60.2 & 76.0 & 74.4 & 70.4 \\
    & DROP {\tiny (F1)} & 77.8 & 76.9 & 88.6 & 84.8 & 73.3 & 91.8 & 87.6 & 87.0 \\
    & GPQA Diamond {\tiny (Pass@1)} & 47.2 & 49.9 & 54.5 & 59.4 & 32.6 & 56.1 & 50.1 & 52.6 \\
    & SimpleQA {\tiny (Correct)} & 6.3 & 9.6 & 6.7 & 12.1 & 12.1 & 24.6 & 38.8 & 9.3 \\ 
    & \cellcolor[HTML]{D7E8E8} Average & \cellcolor[HTML]{D7E8E8}61.2 & \cellcolor[HTML]{D7E8E8}62.9 & \cellcolor[HTML]{D7E8E8}64.5 & \cellcolor[HTML]{D7E8E8}66.9 & \cellcolor[HTML]{D7E8E8}55.8 & \cellcolor[HTML]{D7E8E8}70.9 & \cellcolor[HTML]{D7E8E8}70.8 & \cellcolor[HTML]{D7E8E8}64.4 \\
 \midrule
\multirow{4}{*}{Code} & HumanEval {\tiny (Pass@1)} & 87.8 & 84.8 & 90.2 & 90.2 & 79.3 & 91.5 & 92.1 & 88.4 \\
 & MBPP+ {\tiny (Pass@1)} & 74.6 & 74.6 & 74.9 & 75.9 & 67.2 & 75.7 & 75.9 & 74.5 \\
 & LiveCodeBench {\tiny (Pass@1)} & 31.2 & 32.7 & 36.7 & 43.5 & 22.0 & 42.7 & 38.1 & 32.0 \\
     & \cellcolor[HTML]{D7E8E8} Average & \cellcolor[HTML]{D7E8E8}64.5 & \cellcolor[HTML]{D7E8E8}64.0 & \cellcolor[HTML]{D7E8E8}67.3 & \cellcolor[HTML]{D7E8E8}69.9 & \cellcolor[HTML]{D7E8E8}56.2 & \cellcolor[HTML]{D7E8E8}70.0 & \cellcolor[HTML]{D7E8E8}68.7 & \cellcolor[HTML]{D7E8E8}65.0 \\
  \midrule
\multirow{5}{*}{Math} 
 & MATH {\tiny (EM)} & 82.9 & 83.6 & 86.2 & 88.6 & 57.1 & 89.7 & 79.4 & 85.0 \\
 & AIME24 {\tiny (Pass@1)} & 16.5 & 18.8 & 27.1 & 37.5 & 2.9 & 34.0 & 13.3 & 33.1 \\
 & MATH500 {\tiny (Pass@1)} & 81.9 & 83.1 & 84.8 & 88.1 & 56.2 & 88.9 & 78.4 & 84.8 \\
 & CNMO24 {\tiny (Pass@1)} & 21.1 & 21.9 & 26.0 & 33.2 & 4.3 & 33.9 & 15.6 & 40.6 \\
      & \cellcolor[HTML]{D7E8E8} Average & \cellcolor[HTML]{D7E8E8}50.6 & \cellcolor[HTML]{D7E8E8}51.9 & \cellcolor[HTML]{D7E8E8}56.0 & \cellcolor[HTML]{D7E8E8}61.9 & \cellcolor[HTML]{D7E8E8}30.1 & \cellcolor[HTML]{D7E8E8}61.6 & \cellcolor[HTML]{D7E8E8}46.7 & \cellcolor[HTML]{D7E8E8}60.9 \\
  \midrule
\multirow{4}{*}{Chinese} & CLUEWSC {\tiny (EM)} & 91.7 & 93.1 & 91.6 & 91.4 & 93.0 & 93.5 & 90.3 & 92.6 \\
 & C-Eval {\tiny (EM)} & 87.4 & 88.5 & 82.8 & 84.4 & 78.4 & 86.3 & 77.8 & 92.2 \\
 & C-SimpleQA {\tiny (Correct)} & 42.3 & 51.4 & 46.6 & 62.0 & 56.5 & 68.9 & 61.4 & 56.7 \\
       & \cellcolor[HTML]{D7E8E8} Average & \cellcolor[HTML]{D7E8E8}73.8 & \cellcolor[HTML]{D7E8E8}77.7 & \cellcolor[HTML]{D7E8E8}73.7 & \cellcolor[HTML]{D7E8E8}79.3 & \cellcolor[HTML]{D7E8E8}76.0 & \cellcolor[HTML]{D7E8E8}82.9 & \cellcolor[HTML]{D7E8E8}76.5 & \cellcolor[HTML]{D7E8E8}80.5 \\
 \midrule
      \multirow{4}{*}{Alignment} & IFEval {\tiny (Prompt Strict)} & 78.9 & 84.8 & 84.1 & 84.1 & 58.4 & 86.1 & 85.2 & 82.1 \\
     & AlpacaEval2 {\tiny (GPT-4o Judge)} & 50.0 & 51.3 & 58.1 & 66.8 & 40.4 & 66.5 & 53.6 & 64.4 \\
     & ArenaHard {\tiny (GPT-4o Judge)} & 76.7 & 85.6 & 91.5 & 95.7 & 41.8 & 92.1 & 85.1 & 87.1 \\
    & \cellcolor[HTML]{D7E8E8} Average & \cellcolor[HTML]{D7E8E8}68.5 & \cellcolor[HTML]{D7E8E8}73.9 & \cellcolor[HTML]{D7E8E8}77.9 & \cellcolor[HTML]{D7E8E8}82.2 & \cellcolor[HTML]{D7E8E8}46.9 & \cellcolor[HTML]{D7E8E8}81.6 & \cellcolor[HTML]{D7E8E8}74.6 & \cellcolor[HTML]{D7E8E8}77.9 \\
    \bottomrule
    \end{tabular} }
\end{table}

\subsection{Performance Analysis}
\label{subsec: post-performance}

\paragraph{English Performance}  
Our evaluation results indicate that \redmoechat{} demonstrates stable and comprehensive performance across multiple general English benchmarks. On question-answering tasks such as MMLU, MMLU-Redux, DROP, and GPQA, \redmoechat{} delivers competitive results compared to the Qwen2.5 / Qwen3 series models, demonstrating solid knowledge foundations and reasoning abilities.

\paragraph{Code Performance}  
The model exhibits competitive coding capabilities, though this aspect is not quite as exceptional as its performance in other areas. Compared to the Qwen2.5 series, its coding ability is on par; however, in comparison with more cutting-edge models like Qwen3 and DeepSeek-V3, there remains room for further improvement.

\paragraph{Math Performance}  
The \redmoechat{} showcases outstanding mathematical reasoning skills. The model achieves a score of 33.1 on AIME24, underscoring its advanced problem-solving abilities in complex mathematics. On MATH500, it attains a score of 84.8, outperforming the Qwen2.5 series and approaching state-of-the-art results. It also achieved a high score of 40.6 on CNMO24.

\paragraph{Chinese Performance}  
The \redmoechat{} demonstrates significant advantages in Chinese language tasks. It achieves a score of 92.6 on CLUEWSC, matching industry-leading performance in Chinese semantic understanding. On C-Eval, it achieves 92.2, surpassing all models including DeepSeek-V3 (86.3), thus demonstrating comprehensive mastery of Chinese knowledge. For C-SimpleQA, it scores 56.7, substantially outperforming Qwen-2.5 72B Instruct, though still behind DeepSeek-V3 (68.9).

\paragraph{Alignment Performance}  
With respect to instruction following and alignment with human preferences, \redmoechat{} demonstrates competitive performance on benchmarks such as IFEval, AlpacaEval2, and ArenaHard. These results indicate that the model can accurately interpret and execute complex instructions while maintaining consistency with human intentions and values.

\paragraph{Overall}
Based on the evaluation results, \redmoechat{} demonstrates exceptional performance across Chinese and English general tasks, mathematical reasoning, code generation, and alignment benchmarks while activating only 14B parameters. The model exhibits strong competitive advantages compared to Qwen2.5-32B-Instruct and Qwen2.5-72B-Instruct, and achieves comparable or superior performance relative to Qwen3-32B in bilingual tasks, mathematical reasoning, and alignment capabilities, establishing itself as a highly efficient open-source solution.

\section{MoE Analysis}
\label{sec:moe_analysis}

We examine the expert load of the \redmoe model on the Pile test set. For a specific domain $D$, with $N_D$ tokens processed from this domain by the MoE, the expert load of an expert $E_i$ is defined as:
\begin{equation}
\label{eq:domainspec}
\text{Expert Load}(E_i, D) = \frac{N_{E_i, D}}{N_D}, \quad \in [0, 1]
\end{equation}
where $N_{E_i, D}$ represents the number of tokens from domain $D$ that are routed to expert $E_i$. The expert load indicates how specialized expert $E_i$ is for domain $D$. A value of 1 means that every token from the domain is routed to the expert, while a value of 0 implies the expert is never used for that domain.

\begin{figure}[t]
    \centering
    \includegraphics[width=0.92\textwidth]{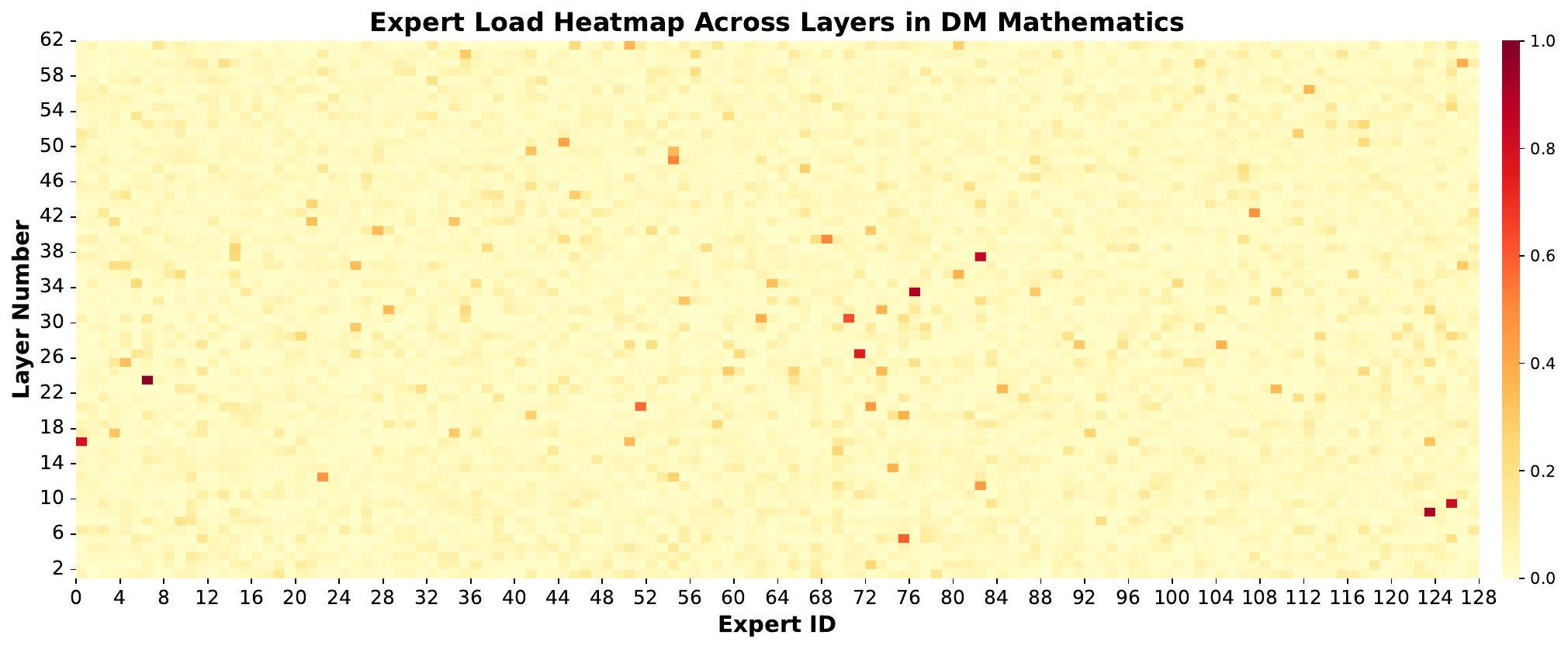}
    \includegraphics[width=0.92\textwidth]{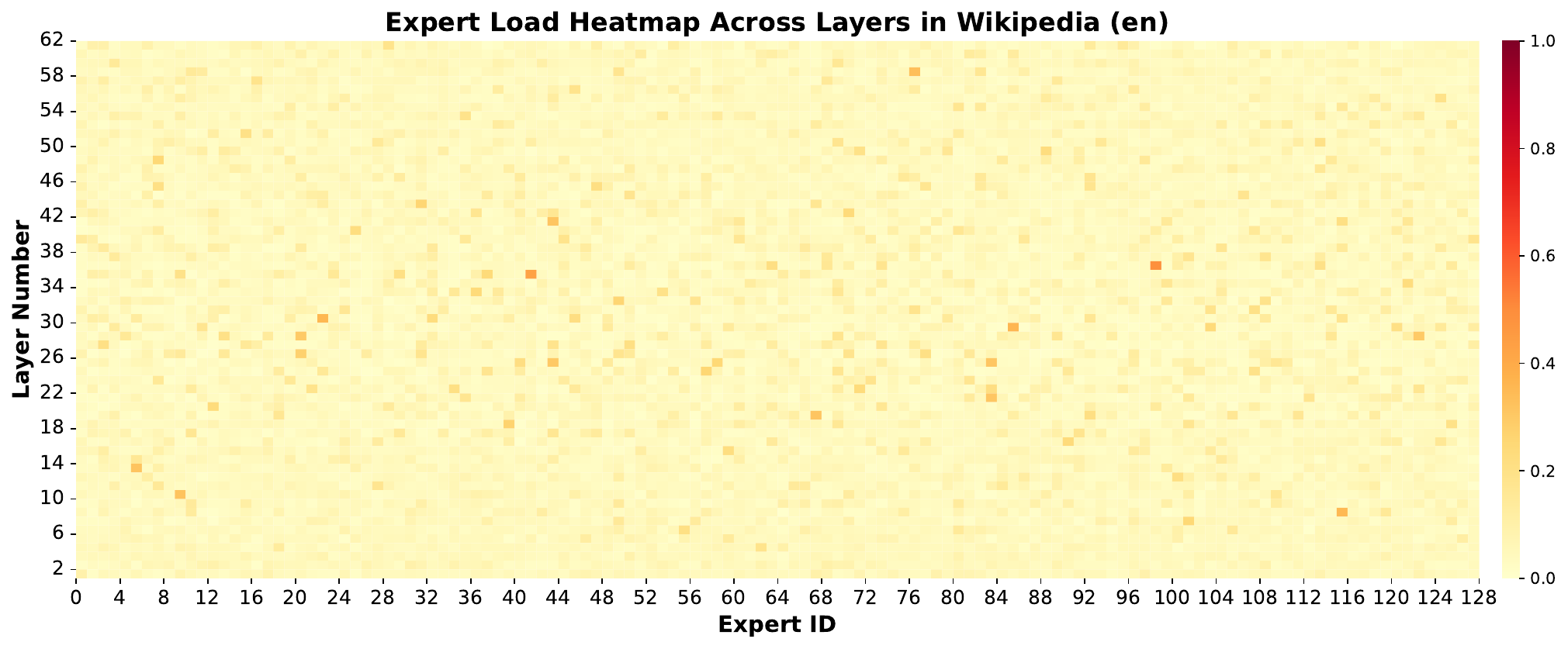}
    \includegraphics[width=0.92\textwidth]{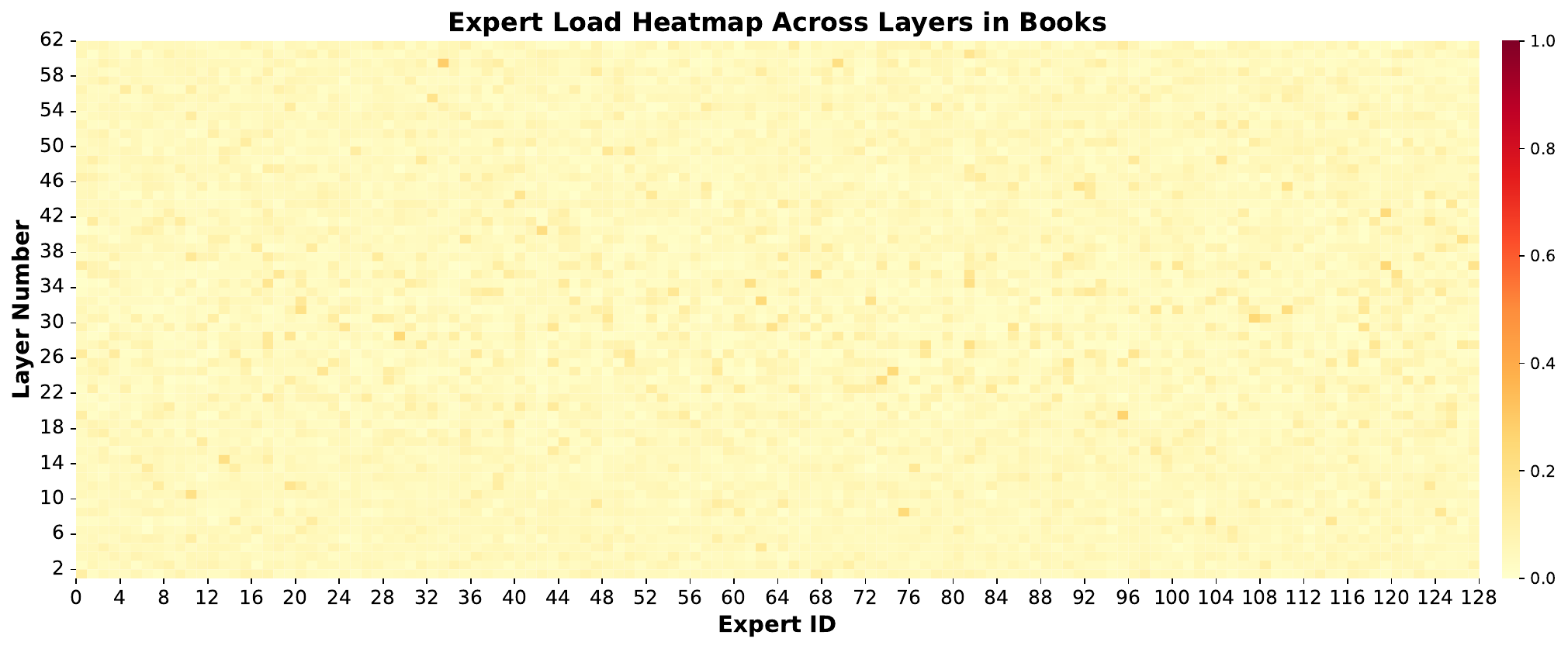}
    \caption{Expert load heatmap across layers on the Pile test dataset.}
    \label{fig:heatmap_1}
\end{figure}

\redmoe shows stronger expert specialization across all layers. In \autoref{fig:heatmap_1}, we observe some experts that are activated significantly more than random selection for specific domains, such as DM Mathematics. In contrast, compared to DM Mathematics, the expert load in Wikipedia appears more balanced. This could be because Wikipedia contains a wide range of knowledge and acts like a knowledge graph encompassing various entities. For the Books domain, which contains various types of data, the expert loads appear more balanced. This highlights that the load balancing function works as intended, enabling the model to effectively utilize all experts for handling generic data.

\section{Conclusion and Future Work}
\label{sec:conclusion}

We introduce \redmoe, a cost-efficient mixture-of-experts model that achieves state-of-the-art performance for its size. By activating only a subset of parameters per token, \redmoe significantly reduces training costs while delivering results comparable with much larger models. Our advanced data processing pipeline produces high-quality training data, and the open-sourcing of intermediate checkpoints provides valuable insights into model learning dynamics. \redmoe demonstrates that efficient design and high-quality data can continually expand the capability boundaries of large language models.

As part of our future work, we aim to train a more powerful model. To achieve an optimal balance between training and inference efficiency, we plan to integrate more efficient architectural designs such as Grouped Query Attention (GQA)~\citep{ainslie2023gqa}, Multi-Head Latent Attention (MLA)~\citep{dsvii}, and Linear Attention~\citep{yang2023gated, katharopoulos2020transformers}. Additionally, we intend to explore the use of more sparse mixture-of-experts (MoE) layers to improve the computational efficiency. Furthermore, since data serves as the foundation of pretraining, we will deepen our understanding of what constitutes optimal training data and explore methods to achieve more human-like learning efficiency, maximizing knowledge acquisition from each training example.





\bibliography{biblio}
\bibliographystyle{colm2024_conference}

\newpage
\appendix
\section*{Appendix}
\section{Authors}



\begin{multicols}{4}
Bi Huo \\
Bin Tu \\ 
Cheng Qin \\ 
Da Zheng \\ 
Debing Zhang \\ 
Dongjie Zhang \\ 
En Li \\ 
Fu Guo \\ 
Jian Yao \\ 
Jie Lou  \\ 
Junfeng Tian \\ 
Li Hu \\ 
Ran Zhu \\ 
Shengdong Chen \\  
Shuo Liu \\ 
Su Guang \\ 
Te Wo \\ 
Weijun Zhang \\ 
Xiaoming Shi \\ 
Xinxin Peng \\ 
Xing Wu \\ 
Yawen Liu \\ 
Yuqiu Ji \\ 
Ze Wen \\ 
Zhenhai Liu \\
Zichao Li \\ 
Zilong Liao 
\end{multicols}

Authors are listed alphabetically by the first name. 
We would like to thank Colin Zhang, Rui Wang, Xing Yu, Di Feng, and Lei Zhang for their support and helpful discussions. We also thank everyone who has contributed to \redmoe{} but is not mentioned in this paper.


\section{Hyperparameters}

We display the pretraining hyperparameter configuration in \autoref{tab:hp} comparing with other relevant models.

\begin{table}[ht]
\centering
\caption{Architectural and pretraining hyper-parameters of \redmoe compared with recent models.
WSD = weight-stable decay~\citep{minicpm}.}
\begin{tabular}{l|ccc|c}
\toprule
& \textbf{Qwen2.5 72B} & \textbf{DeepSeek-V2} & \textbf{DeepSeek-V3} & \textbf{\redmoe} \\
\midrule
Architecture         & Dense   & MoE    & MoE     & MoE        \\
Active parameters    & 72B     & 21B    & 37B     & 14B        \\
Total parameters     & 72B     & 236B   & 671B    & 142B       \\
\midrule
Hidden size          & 8,192   & 5,120  & 7,168   & 4,096      \\
Activation           & SwiGLU  & SwiGLU & SwiGLU  & SwiGLU     \\
FFN size             & 29,568  & 12,288 & 18,432  & 10,944     \\
Vocabulary size      & 152,064 & 102,400& 129,280 & 152,064    \\
Attention type       & GQA     & MLA    & MLA     & MHA        \\
Attention heads      & 64      & 128    & 128     & 32         \\
KV heads             & 8       & 128    & 128     & 32         \\
Layers               & 80      & 60     & 61      & 62         \\
Attention biases     & Yes    & No     & No      & No         \\
QK normalization     & No      & No     & No      & Yes        \\
Positional encoding  & RoPE    & RoPE   & RoPE    & RoPE       \\
MoE FFN size         & –       & 1,536  & 2,048   & 1,408      \\
Routed experts       & –       & 160    & 256     & 128        \\
Shared experts       & –       & 2      & 1       & 2          \\
Top-k experts        & –       & 6      & 8       & 6          \\
Init. std            & –       & 0.006  & 0.006   & 0.006      \\
\midrule
LR schedule          & –       & WSD    & WSD     & WSD        \\
Warmup steps         & –       & 2,000  & 2,000   & 4,000      \\
Peak LR              & –       & 2.4E-4 & 2.2E-4  & 3.0E-4     \\
Optimizer            & –       & AdamW  & AdamW   & AdamW      \\
Weight decay         & –       & 0.1    & 0.1     & 0.1        \\
Beta1                & –       & 0.9    & 0.9     & 0.9        \\
Beta2                & –       & 0.95   & 0.95    & 0.95       \\
\midrule
Pretraining tokens   & 18T     & 8.1T   & 14.8T   & 11.2T      \\
Sequence length  & 4,096   & 4,096  & 4,096   & 8,192      \\
\bottomrule
\end{tabular}
\label{tab:hp}
\end{table}

\section{Web Data Curation}
\label{appsec:data}

\subsection{Document Preparation Phase}

\paragraph{URL Filtering}  Similar to RefineWeb \citep{refineweb}, URL filtering focuses on filtering toxic documents based solely on the URL, rather than the document content. We maintain a blocklist of domains, implementing careful manual verification to ensure removal of pages associated with adult themes, gambling, and other toxic topics.

\paragraph{Text Extraction} Following RefineWeb \citep{refineweb}, we employ \textit{trafilatura}\citep{barbaresi-2021-trafilatura}  to extract the main text from webpages. To mitigate the presence of irrelevant content, we implement a series of custom optimizations for \textit{trafilatura}, including adjustments to HTML patterns, keyword filtering, and content length specifications. 

\paragraph{Language Identification} Leveraging the fastText language identification model from CCNet~\citep{ccnet16}, we classify the language of each document efficiently. To ensure the quality and relevance of the dataset, we discard any documents with a language classification confidence score below a threshold of $0.65$. This step is crucial for filtering out noisy or ambiguous data, particularly in multilingual datasets, and helps maintain a consistent and high-quality corpus.
    
\paragraph{Identity Removal} To ensure unique identification, we compute a 32-character hexadecimal string for each document using the MD5 digest algorithm. Duplicate documents are then removed randomly to retain only unique content for further processing. Prior to deduplication, we preprocess the text by removing all punctuation, normalizing characters to the NFD Unicode format\footnote{\url{https://en.wikipedia.org/wiki/Unicode_equivalence}}, converting text to lowercase, and eliminating extra spaces between words. This step is essential for preserving dataset integrity and ensuring that only distinct documents are included in the final collection.

\subsection{Rule-Based Processing Phase}

\paragraph{Line-wise Inter-Document Deduplication} We introduce a line-level deduplication method specifically designed to target repetitive patterns commonly found in the head and tail sections of web documents, such as advertisements, navigation bars, and other non-informative content. This approach effectively reduces inter-document redundancy while preserving the essential and meaningful content, ensuring a cleaner and more focused dataset. First, we extract the first $5$ lines and the last $5$ lines of each document, splitting them into individual lines while discarding any lines that are empty or contain only whitespace or symbols. Next, we calculate the frequency of each line across the entire dataset. For any line that appears more than 200 times, we retain only its first 200 occurrences within the dataset and remove all additional instances from their respective documents. If a document does not contain any lines that need to be removed, it is preserved in its original form. Otherwise, the specified lines are removed, and the remaining parts are concatenated to form the final document. This method is implemented in an incremental manner and within distributed systems, making it more efficient and scalable for large-scale datasets.


\paragraph{Rule-Based Filtering} We develop a filtering system based on RefinedWeb~\citep{refineweb} and Gopher~\citep{gopher}, combining precise heuristic rules and statistical features to systematically remove low-quality content. Our pipeline includes empty content removal, advertisement and registration prompt filtering, domain/URL/title-based meta filtering, wiki/code diff detection, structural anomaly and duplicate content elimination, and content quality filtering. Additionally, we perform manual review and tailored cleaning for major domains, focusing especially on the top 1,000 sites that comprise 60\% of the data, to ensure consistently high data quality.

\paragraph{Fuzzy Deduplication} We employ MinHash \citep{broder1997resemblance} and Locality-Sensitive Hashing (LSH) \citep{leskovec2020mining} to perform approximate deduplication. The process involves the following steps: First, we apply the same text standardization as in the Identity Removal step. Next, we tokenize the text using the Jieba tokenizer \footnote{\url{https://github.com/fxsjy/jieba}} to handle Chinese text, followed by 5-gram processing. Using the resulting 5-gram tokens, we compute 2048 MinHash values for each text. These MinHash values are then divided into 128 bands, each containing 16 rows, as part of the LSH configuration. For any samples that collided within the same band, we retain only one instance. This approach enables us to efficiently deduplicate text pairs with a Jaccard similarity of 80\%, achieving a high probability of 97.42\%.

\subsection{Model-Based Processing Phase}

\paragraph{Web-Type Classification Model} Similar to how humans perceive and categorize information, we implement a web content classifier that distinguishes various types of web pages. The classifier utilizes a 1.5B model that categorizes content into text-rich detail pages (such as articles) and non-essential web pages, including tool pages (e.g., maps, public transit route queries), audio pages, video pages, forum sites, and adult websites.
Through this process, we selectively preserve only high-quality detail pages for subsequent processing.


\paragraph{Web Clutter Removal Model} We implement a line-wise web clutter removal model to eliminate unnecessary web page elements such as borders, advertisements, navigational components, and repetitive content, focusing exclusively on main content. This model evaluates each individual line based on its informational value, relevance, and quality, assigning a score from 0 to 1. We fine-tune a 1.5B model and find that it significantly improves the overall data quality.

\paragraph{Quality Model} The quality model performs comprehensive multidimensional analysis to evaluate and score training samples~\citep{qwen2_5, penedo2024the}.
We design a comprehensive annotation specification to evaluate text quality, including text fluency, coherence, information redundancy, and repetition patterns.
In order to balance efficiency and model performance, we adopt a k-fold cross validation method with a 1.5B model and evaluate its discriminative ability using the area under the ROC curve (AUC).
Using the quality score generated by this model, we set a threshold to only retain high-quality data from the corpus.

\paragraph{Semantic Deduplication} In line with \cite{llama3}, we implement a semantic deduplication strategy to eliminate documents with high semantic overlap. Initially, we employ \texttt{BGE-M3} \citep{bge-m3} as an embedding model to create embeddings for each document. Subsequently, we apply the KMeans algorithm \citep{macqueen1967some} to categorize these embeddings into clusters. 
This clustering step significantly reduces computational costs in subsequent pairwise similarity calculations. Following the clustering phase, we compute the pairwise cosine similarity between document embeddings within the same cluster. Duplicates are identified based on a predefined similarity threshold, which is set to $0.95$ in our process. By doing so, we ensure the preservation of varied and distinctive content while discarding redundant semantic information.

\paragraph{Category Balancing}
To curate a more balanced and information-rich training dataset, we first develop a fine-grained, 200-class content classifier to categorize the vast web data. This classifier enables us to identify and quantify the distribution of various content types within the corpus. Based on this categorization, we proportionally increase the representation of knowledge-based and factual content, such as encyclopedia entries, educational materials, and popular science articles. Meanwhile, we deliberately reduce the proportion of less informative or structurally formulaic content, including fiction (e.g., science fiction novels) and product descriptions. This balancing process ensures that the training dataset is both diverse and skewed towards sources that are rich in reliable information, thereby fostering the development of language models with stronger comprehension, reasoning, and knowledge retention abilities.



\end{document}